\def\year{2022}\relax
\documentclass[letterpaper]{article} 
\usepackage{aaai22}  
\usepackage{times}  
\usepackage{helvet}  
\usepackage{courier}  
\usepackage[hyphens]{url}  
\usepackage{graphicx} 
\usepackage{natbib}  
\usepackage{caption} 
\DeclareCaptionStyle{ruled}{labelfont=normalfont,labelsep=colon,strut=off} 
\usepackage{algorithm}
\usepackage{algorithmic}
\usepackage{amsmath}
\usepackage{subfig}
\usepackage{amssymb}
\usepackage{marvosym}
\usepackage{hyperref}
\usepackage{newfloat}
\usepackage{listings}
\floatstyle{ruled}
\newfloat{listing}{tb}{lst}{}
\floatname{listing}{Listing}
\title{Improving Human-Object Interaction Detection via\\
Phrase Learning and Label Composition}
\author{
    {Zhimin Li\textsuperscript{\rm 1}\thanks{Equal contribution. This work was done when Zhimin Li was an intern at Megvii.}} \qquad Cheng Zou\textsuperscript{\rm 2}{$^{*}$}\textsuperscript{\Letter} \qquad Yu Zhao\textsuperscript{\rm 2} \qquad Boxun Li\textsuperscript{\rm 2} \qquad Sheng Zhong\textsuperscript{\rm 1}\textsuperscript{\Letter}\\
}
\affiliations {
    \textsuperscript{\rm 1} National Key Laboratory of Science and Technology on Multispectral Information Processing,
School of Artificial Intelligence and Automation, Huazhong University of Science and Technology\\
    \textsuperscript{\rm 2} Megvii Technology\\
    \{lizm, zhongsheng\}@hust.edu.cn \quad \{zoucheng, zhaoyu03, liboxun\}@megvii.com
}

\usepackage{bibentry}

\begin{document}
\maketitle

\begin{abstract}
Human-Object Interaction (HOI) detection is a fundamental task in high-level human-centric scene understanding. We propose PhraseHOI, containing a HOI branch and a novel phrase branch, to leverage language prior and improve relation expression. Specifically, the phrase branch is supervised by semantic embeddings, whose ground truths are automatically converted from the original HOI annotations without extra human efforts. Meanwhile, a novel label composition method is proposed to deal with the long-tailed problem in HOI, which composites novel phrase labels by semantic neighbors. Further, to optimize the phrase branch, a loss composed of a distilling loss and a balanced triplet loss is proposed. Extensive experiments are conducted to prove the effectiveness of the proposed PhraseHOI, which achieves significant improvement over the baseline and surpasses previous state-of-the-art methods on Full and NonRare on the challenging HICO-DET benchmark. 

\end{abstract}
\section{Introduction}

Human is the main focus in visual world\cite{gupta2016cross}. Figuring out the relationships between human and its context is of great importance. Human-Object Interaction (HOI) is to detect interactive relation between human and the surroundings. HOI detection plays an important role in high-level human-centric scene understanding, and has attracted considerable research interest recently~\cite{ulutan2020vsgnet, eccv2020actioncoprior, liao2020ppdm, wang2020irnet,eccv2020actioncoprior,eccv2020keycues, bansal2020detecting, gao2018ican}, resulting in significant improvement~\cite{eccv2020dualgraph, eccv2020visualcomlear, li2020pastanet, zou2021_hoitrans}.

\begin{figure}[t]
    \centering
    \includegraphics[scale=0.39]{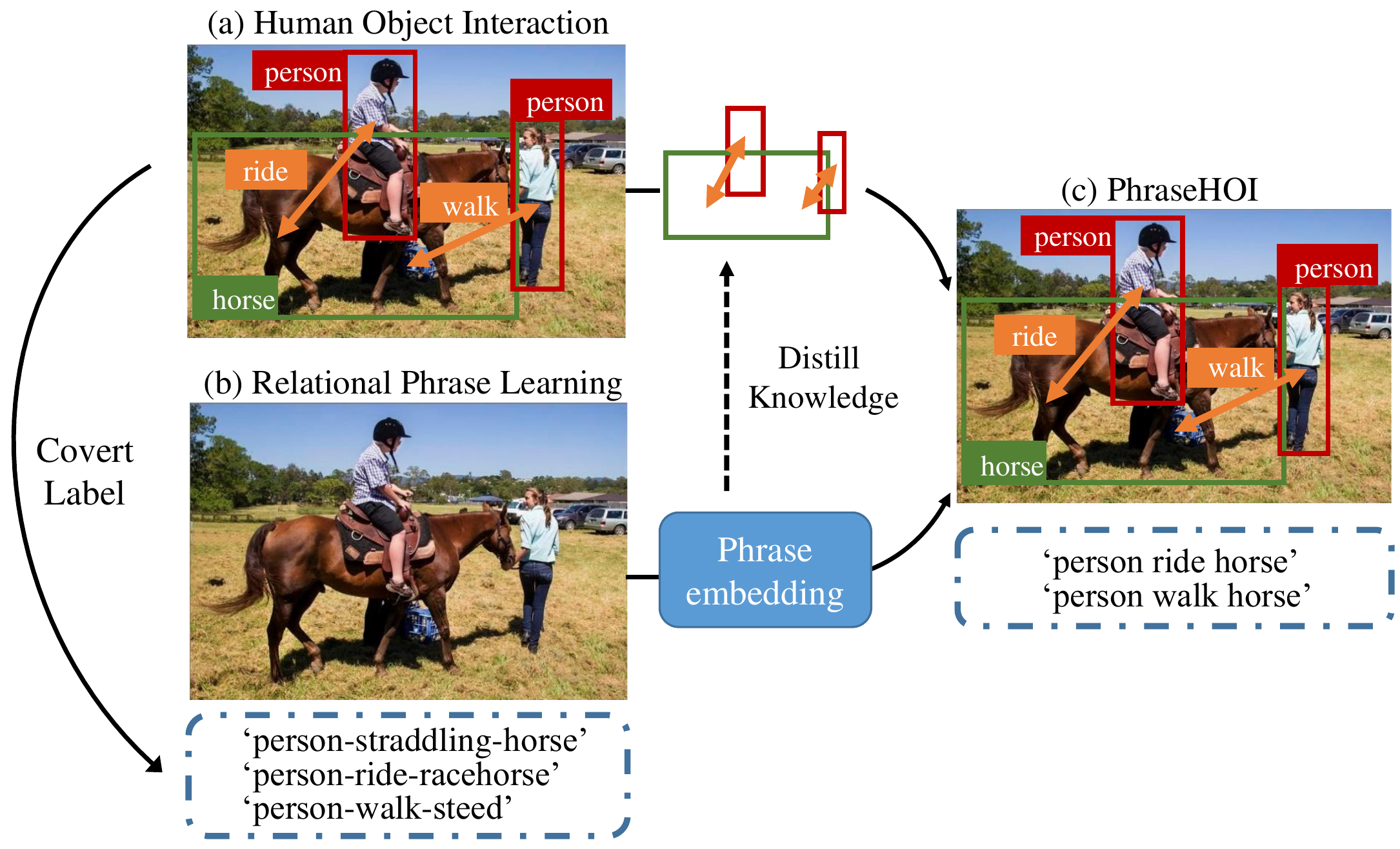}
    \caption{(a) HOI task, which inputs an image and outputs human bounding boxes, object bounding boxes and the interactions between them. (b) Relational phrase learning task, which inputs an image and outputs several natural language phrases to describe the interactions between humans and objects. (c) Proposed PhraseHOI, which inputs an image and outputs all of the above.}
    \label{fig:multi-task}

\end{figure}

The task of HOI detection aims at localizing and recognizing relationships between humans and objects. Fig.~\ref{fig:multi-task}(a) illustrates an example of HOI, which inputs an image and outputs human bounding boxes, object bounding boxes and the interaction categories between them.

It's hard to obtain labeled data for HOI due to the various interactions between humans and objects in reality. So it's natural to think about using data from other sources as an alternative way to improve the performance of HOI detection. In this paper, we propose a novel relational phrase learning task derived from HOI to utilize language knowledge. 
Fig.~\ref{fig:multi-task}(b) illustrates an example of relational phrase learning. It inputs an image and outputs several natural language phrases to describe the interactions between humans and objects. Here we call these phrases `relational phrases', and the ground truths for them are automatically converted from the original HOI annotations without extra human efforts. 

Once staring at Fig.~\ref{fig:multi-task}(a) and Fig.~\ref{fig:multi-task}(b) together, we find that these two tasks both model the interactions between humans and objects, although one is from visual modality and the other is from language modality. We hypothesize that these two tasks share a common part in there knowledge space, and joint optimization may help to distill knowledge from the relational phrase learning task to HOI task, which will improve the performance of HOI detection. Specifically, influenced by the task of cross-modal knowledge distillation~\cite{gupta2016cross}, our knowledge comes from the teacher, a pretraind word2vec model, then to teach 
the HOI model.
Also, HOI detection is confronted with a long-tailed problem due to the large enumeration space of verb-object pairs. This is really a challenge because there exists very few samples for some rare HOI categories. To recognize something unknown, we observe that human beings tend to recall information from those related or similar things they have already known.
For instance, the scene of `person ride giraffe' is rare, however human have learned the prior knowledge of `person ride horse' and `person watch giraffe', then we could quickly compose `person ride giraffe' with the prior knowledge and remember it. Inspired by this, we construct semantic neighborhood set for each word and then propose a novel label composition method to extend label space. Specifically, we composite new labels for a given relational phrase by replacing a verb word or an object word according to their semantic neighbors. This composition method is quite different from previous work~\cite{bansal2020detecting, eccv2020visualcomlear}, all of which do augmentation in input feature space and need to break the pipeline, while ours is in the label space. 

In this paper, we propose a multi-task architecture named PhraseHOI to jointly optimize the related tasks, i.e. HOI and relational phrase learning, resulting in significant performance improvement on HOI. It consists of a transformer backbone and two parallel branches. One is HOI branch, and its design follows~\cite{zou2021_hoitrans}. The other is phrase branch, it outputs an embedding which can be translated into a relational phrase through Look-Up-Table (LUT). Fig.~\ref{fig:multi-task}(c) illustrates the concept of the proposed method, which inputs an image and outputs human bounding boxes, object bounding boxes, interaction categories, and relational phrases. To optimize the phrase branch, a distilling loss along with triplet loss is proposed, to globally keep the feature distribution of the original language space, and to locally make semantically similar embeddings discriminative to each other.

\par
The contributions are summarized as follows,
\begin{itemize}
\item  As far as we know, we are the first to creatively use HOI annotations as phrases. And based on this, we propose a novel relational phrase learning task derived from HOI to describe interactions between humans and objects with natural language phrases.
\item We propose PhraseHOI to jointly optimize HOI detection and relational phrase learning to improve the performance of HOI.
\item We propose a novel label composition method to address the long-tailed problem in HOI, which composites new samples by semantic neighbors in label space.
\item Our method gets significant improvement compared to our baseline, set state-of-the-art on Full and NonRare on the challenging HICO-DET benchmark.
\end{itemize}

\section{Related Work}

\textbf{External Knowledge. }Using knowledge of different types to improve HOI detection is becoming popular, because these extra knowledge brings in richer information. In~\cite{li2020pastanet, li2019tin, shen2018scaling, wan2019pose, zhou2019relation, zhong2020polysemy, fang2018pairwise}, human pose or body part is introduced to pay more attention to local area rather than treating the whole human area uniformly. In~\cite{xu2019learning, eccv2020dualgraph, eccv2020keycues, eccv2020actioncoprior, bansal2020detecting}, word embedding or language prior is introduced, e.g. Xu~\cite{xu2019learning} used GloVe~\cite{pennington2014glove} to generate word embedding as part of graph node representation. In this work, we use language prior to enhance HOI detection, but unlike other methods, the language embedding in our work is the output and it is the target for the model to learn, rather than the input feature to the model. This makes the 
model flexible and can be regarded as a meta-architecture for other HOI detectors.

\begin{figure*}
	\begin{center}
		\includegraphics[width=0.95\linewidth]{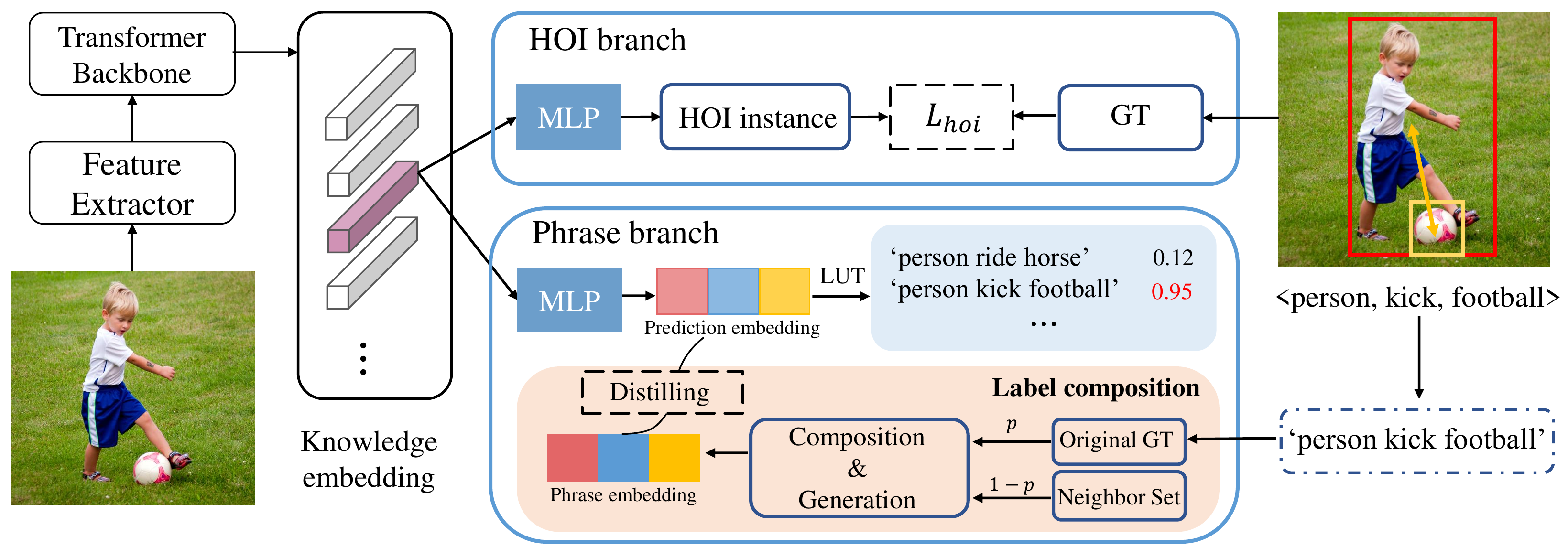}
	\end{center}
	\caption{Overall architecture. The proposed PhraseHOI has a transformer backbone and two parallel branches, one is HOI branch and the other is phrase branch. The training and inference of HOI branch follows~\cite{zou2021_hoitrans}. The phrase branch first predicts phrase embedding and then translates it into a relational phrase through Look-Up-Tab~(LUT). In training, the original HOI triplet is first converted to a relational phrase, and then sent to a label composition module to generate new phrases, after that, the phrase is encoded to a phrase embedding as GT to supervise the prediction embedding.}
	\label{fig:framework}

\end{figure*}

\textbf{Long-tailed Problem. }HOI detection is facing a long-tailed distribution problem because it is difficult to collect labeled training data for all HOI categories, due to the large enumeration space of verb-object pairs. Shen~\cite{shen2018scaling} may be the first one to introduce zero-shot learning approach to tackle the long-tailed problem in HOI by disentangling reasoning on verbs and objects. Bansal~\cite{bansal2020detecting} incorporates the idea that human interacting with functionally similar objects look similar via a functional generalization module, resulting in feature augmentation during training according to the semantic representation similarities of objects. VCL~\cite{eccv2020visualcomlear} addresses the long-tailed issues via visual composition learning, which composes new interaction samples in the visual feature space by decomposing an HOI representation into object and verb specific features. These two methods are doing augmentation in input feature space, while in this work, we propose a novel label composition method to augment in the label space, which is quite different from~\cite{bansal2020detecting, eccv2020visualcomlear}.


\textbf{Context Modeling. }It's vitally important to model the context information properly for HOI detection. In~\cite{gao2018ican, li2019tin, wang2019deep, ulutan2020vsgnet}, a two-channel binary image representation is used to encode the spatial context. In~\cite{eccv2020keycues}, image scene class information is introduced to encode global visual context. DRG~\cite{eccv2020dualgraph} realized the problem that most methods ignore the contextual information from other interactions in the scene and predict the current interaction between each human-object pair in isolation, so a dual graph is proposed to enable knowledge transfer among objects. CHG~\cite{eccv2020hetegraph} found that relations between homogeneous entities and heterogeneous entities should not be equally the same, so a contextual heterogeneous graph is built to model the intra-class context and inter-class context, which is more elaborate than ~\cite{zhou2019relation}. Most recently, ~\cite{zou2021_hoitrans, qpic2021, ASNET, CDN} reasons about the interactions between humans and objects from global image context with transformer. In this work, we introduce transformer to model the context information following~\cite{zou2021_hoitrans} due to its large receptive field and simple architecture.

\section{Methodology}

\subsection{Overview}
In this section, we first propose a multi-task architecture named PhraseHOI to jointly optimize the related tasks, i.e. HOI and relational phrase learning, in a unified network. Second, to learn the relational phrase task, we creatively generate ground truths from original HOI annotations. Then, a novel label composition method is proposed to handle the long-tailed problem in HOI. After that, it's the phrase embedding learning loss. Finally, we introduce two available post processing methods.

\subsection{Network Architecture}
The proposed PhraseHOI has a transformer backbone and two parallel branches, one is HOI branch and the other is phrase branch. The HOI branch outputs HOI instance, i.e. a quintuple of (human class, interaction class, object class, human box, object box) and the phrase branch outputs a phrase embedding  which  can  be  translated to a relational phrase through Look-Up-Table (LUT).


\textbf{Transformer Backbone. }The network design follows~\cite{zou2021_hoitrans} because transformer provides large receptive field and rich context information, both locally and globally, which is important for HOI detection. Given an input image, the transformer backbone outputs $N$ feature embeddings (denoted as knowledge embedding in Fig.~\ref{fig:framework}), which have encoded the relationship knowledge between humans and objects implicitly.

\textbf{HOI Branch. }This branch is attached to the knowledge embedding, and consists of five MLP heads which directly predict the HOI instance, i.e. a quintuple of (human class, interaction class, object class, human box, object box). The training loss and training target for this branch follows~\cite{zou2021_hoitrans}.

\textbf{Phrase Branch. }The phrase branch lies in parallel to the HOI branch, and both of them are attached to the same knowledge embedding, implying that these two branches share the same knowledge space. As a result, there is always a phrase embedding counterpart associated with an HOI instance in the output. 
After a three-layer MLP, the knowledge embedding is converted to a phrase embedding, which is a 900-d vector, and can be translated to a relational phrase through LUT. In training, the original relational phrase is first sent to a label composition module to composite a new phrase. Then the composed phrase is sent to an embedding generation module to generate a 900-d phrase embedding, which is the training target for the phrase branch. The model is jointly optimized and the ground truth for training the phrase branch is detailed in the following section. Fig.~\ref{fig:framework} illustrates the overall architecture.

\subsection{Relational Phrase Generation}
\label{section:ground_truth}
To train the novel phrase branch, we creatively construct a relational phrase target from a given original HOI annotation. Specifically, we directly use the `word' of human, verb and object to construct a three-word phrase, e.g. the phrase `human ride horse' can be derived from the original HOI triplet (human, ride, horse). After that, each of the words is sent to a pretrained word embedding model, e.g. word2vec~\cite{mikolov2013distributed}, to get an embedding. Finally, the three embeddings are concatenated in 
`human-verb-object' order to form a phrase embedding, and this is the ground truth for the phrase branch to learn. 

\begin{figure}[t]
	\begin{center}
		\includegraphics[width=1.0\linewidth]{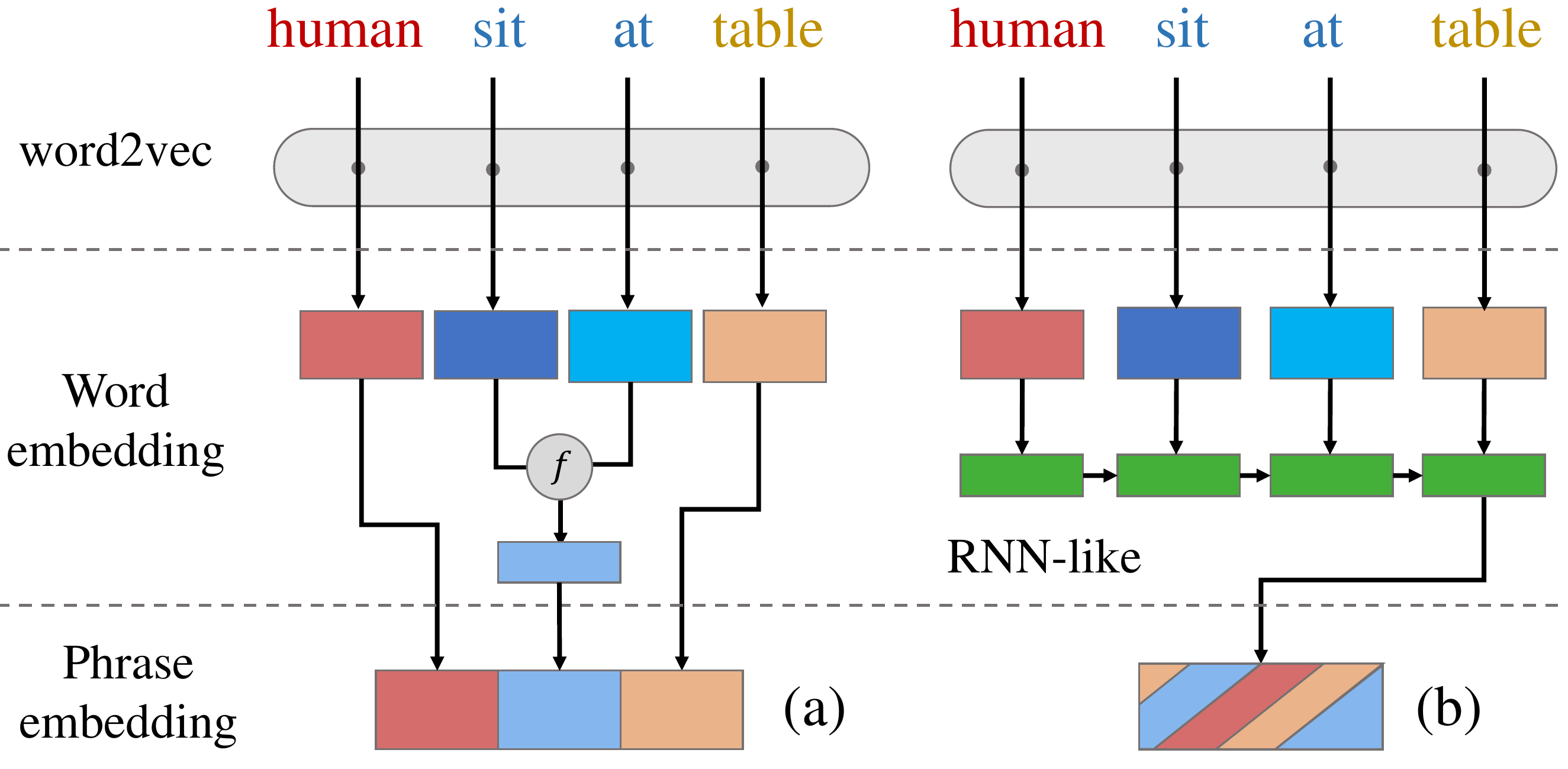}
	\end{center}
	\caption{Illustration of two methods for phrase embedding generation. (a) the concatenated method. (b) the language model encoded method.}
	\label{fig:embedding_generation}
\end{figure}

It's noted that some compound words can not be found in word embedding model, e.g. `dining table'. In this case we split it at first and then get the mean embedding of them. Fig.~\ref{fig:embedding_generation}(a) illustrates this procedure. And there is another reasonable embedding generation method shown in Fig.~\ref{fig:embedding_generation}(b), feeding all the words to a time serial model to get the output hidden embedding, which will be discussed in experiments.

\subsection{Label Composition}
\label{section:labe_composition}
\begin{figure}[t]
	\begin{center}
		\includegraphics[width=1.0\linewidth]{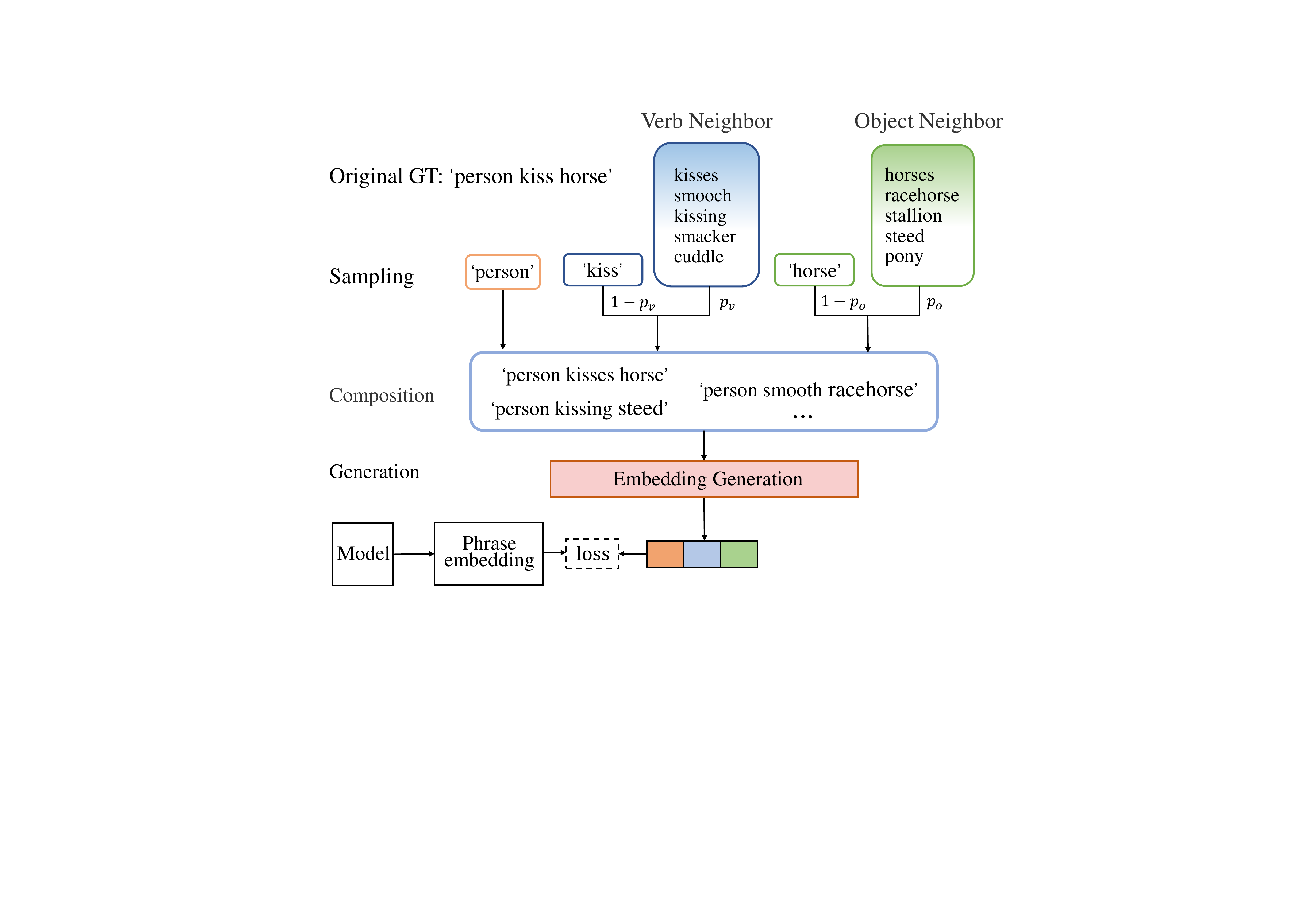}
	\end{center}
	\caption{Illustration of the label composition method. Given a GT phrase in the format of `human verb object', we sample from the verb word's neighborhood set and the object word's neighborhood set to composite new phrases.}
	\label{fig:label_composition}
\end{figure}

HOI detection is facing a long-tailed problem due to the lack of training data on rare HOI categories. Recent studies~\cite{bansal2020detecting, eccv2020visualcomlear} indicates that augmentation strategies do help. According to~\cite{lu2016visual}, there is a good property in language feature space that semantically similar objects are close to each other. Based on this, we propose a novel label composition method to tackle the long-tailed problem in HOI, as shown in Fig.~\ref{fig:label_composition}.

It is noteworthy that the proposed method is quite different from~\cite{eccv2020visualcomlear} and~\cite{bansal2020detecting}. Both of them do augmentation in input feature space, which usually needs a multi-stage pipeline, while ours is in label space, nearly all kinds of pipelines fit, two-stage, one-stage, and end-to-end.

To make it simple, let's take HICO-DET dataset which contains 117 different verbs and 80 different objects as an example. For each verb word, we find its top $K$ nearest neighbors in the whole space, and only keep those words whose similarities are larger than a given similarity threshold $t_{sim}$. Thus, each verb word will have its own neighborhood set. For example, given query word `kiss' and similarity threshold $t_{sim}=0.7$, we get its neighborhood set \{(`kisses', 0.81), (`smooch', 0.76), ('kissing', 0.71)\}, where the float number in the bracket indicates the similarity with query word `kiss'. It's the same with each object word.

In training, given a ground truth, e.g. `human kiss horse', then for verb `kiss', we sample from its neighborhood set \{`kisses', `smooch', `kissing'\} at a probability of $p_v$ and $1-p_v$ to keep itself, for object `horse', we sample from its neighborhood set \{`horses', `racehorse', `stallion', `steed'\} at a probability of $p_o$ and $1-p_o$ to keep unchanged. In this way, the original one ground truth `human kiss horse' is expanded to 20 different samples, which dramatically increase the label diversity and is proved to be effective.

This augmentation makes sense because the neighborhood set for each query word is carefully chosen by their similarities in the whole space, i.e. only top closest neighbors are taken into consideration. In other words, a word is replaced by its semantically similar neighbors during new phrase composition.


\subsection{Distill with Triplet Loss}
\label{section:loss}

\begin{table*}[t]
\setlength{\tabcolsep}{3pt}
\begin{tabular}{lcc|ccc|ccc}
\hline \hline
\multicolumn{1}{l}{} & \multicolumn{1}{l}{} & \multicolumn{1}{l}{} & \multicolumn{3}{|c}{Default}    & \multicolumn{3}{|c}{Known Object}                     \\
Methods        &  Backbone                  & Knowledge        & Full$\uparrow$                  & Rare$\uparrow$                  & NRare$\uparrow$               & Full$\uparrow$ & Rare$\uparrow$ & NRare$\uparrow$ \\ \hline
 \textit{Two-stage methods} & & & & & & & & \\

CHG~\cite{eccv2020hetegraph}             & ResNet-50                          &                                           & 17.57                           & 16.85                           & 17.78                           &  21.00              &    20.74            &    21.08               \\
VSGNet~\cite{ulutan2020vsgnet}           & ResNet152                           &                                            & 19.80                           & 16.05                           & 20.91                           &     -           &     -           &      -             \\
FCMNet~\cite{eccv2020keycues}            & ResNet-50                           & \checkmark                 & 20.41                           & 17.34                           & 21.56                           &   22.04             &       18.97         &    23.12               \\
ACP~\cite{eccv2020actioncoprior}         & ResNet-152                         & \checkmark                & 20.59                           & 15.92                           & 21.98                           &        -        &    -            &    -               \\
Bansal et al.~\cite{bansal2020detecting} & ResNet-50-FPN                   &                             & 21.96                           & 16.43                           & 23.62                           &      -          &           -     &     -              \\
PD-Net~\cite{zhong2020polysemy}           & ResNet-152                        &                             & 20.81                           & 15.90                         & 22.28                          &      24.78          &   18.88             &       26.54            \\
PastaNet~\cite{li2020pastanet}           & ResNet-50                          & \checkmark                 & 22.65                           & 21.17 & 23.09                           &       24.53         &       23.00         &     24.99              \\
VCL~\cite{eccv2020visualcomlear}         & ResNet101                      &                                           & 23.63                           & 17.21                           & 25.55                           &     25.98           &    19.12            &  28.03                 \\
DRG~\cite{eccv2020dualgraph}             & ResNet-50-FPN                   &                              & 24.53                           & 19.47                           & 26.04                           &     27.98           &   23.11           &   29.43                \\  \hline \hline
\textit{One-stage methods}& & & & & & & & \\
UnionDet~\cite{eccv2020uniondet}         & ResNet-50-FPN                  &                                           & 17.58                           & 11.52                           & 19.33                           &  19.76              &  14.68              &     21.27              \\ 
IPNet~\cite{wang2020irnet}               & Hourglass                            &                                         & 19.56                           & 12.79                           & 21.58                           &    22.05            &      15.77          &     23.92              \\
PPDM~\cite{liao2020ppdm}                 & Hourglass                        &                                          & 21.73                          & 13.78                           & 24.10                           &  24.58              &   16.65             &           26.84       \\ 
AS-Net~\cite{ASNET}                      & ResNet-50                        &                                          & 28.87                         & \textbf{24.25}                           & 30.25                           &  31.74              &   27.07             &           33.14        \\
\hline \hline
\textit{End-to-End methods} & & & & & & & &\\    

HOTR~\cite{HOTR}                      & ResNet-50                        &                                          & 25.10                          & 17.34                           & 27.42                           & -              &   -            &         -  \\
QPIC~\cite{qpic2021}                      & ResNet-50                        &                                          & 29.07                          & 21.85                           & 31.23                         & 31.68              &  24.14            &       33.93\\
\hline
HOI Transformer~\cite{zou2021_hoitrans}                                 & ResNet-50                               &                                           & 23.46                          & 16.91                           & 25.41                          &   26.15             &       19.24         &      28.22             \\
HOI Transformer~\cite{zou2021_hoitrans}       & ResNet-101                            &                                          & 26.61                          & 19.15                           & 28.84                        &   29.13             &   20.98            &     31.57      \\
\textbf{Ours}                                 & ResNet-50                              &                         \checkmark                 & 29.29                       & 22.03                       & 31.46                     &    31.97 & 23.99             &  34.36             \\
\textbf{Ours}                                 & ResNet-101                            &                          \checkmark                   & \textbf{30.03}                          & 23.48                        & \textbf{31.99}                         &    \textbf{33.74}           &  \textbf{27.35}         &       \textbf{35.64}         \\

\hline \hline
\end{tabular}
\caption{Comparison to state-of-the-art methods on HICO-DET test set. Knowledge means the methods use pose or language knowledge. Our method achieves the best performance on Full and NonRare categories. NRare is short for NonRare.}
\label{tab:hico}
\end{table*}

We design the following loss to make sure that the multi-branch can be jointly trained:
\begin{equation}
    \mathcal{L}_{total} = \mathcal{L}_{hoi} + \alpha *  \mathcal{L}_{phrase}
    \label{eq:loss_total}
\end{equation}
where $\mathcal{L}_{hoi}$ is the loss for HOI branch, which consists of the bounding box regression loss for human and object boxes, the classification loss for human, object and interaction categories. The implementation details of $\mathcal{L}_{hoi}$ follows ~\cite{zou2021_hoitrans}. $\mathcal{L}_{phrase}$ implies the phrase embedding learning loss for phrase branch, and $\alpha$ is used to balance these two tasks in training. 

Intuitively, $\mathcal{L}_{phrase}$ should be a distilling loss, since the relational phrase learning can be interpreted as a knowledge distilling procedure, and it aims to predict a phrase embedding which is close to the given ground truth. However, this is not enough. Distilling only constrains the relationship between predictions and ground truth, it is trying to pull the faraway predictions close to the ground truth, but it does not consider the relationship between predictions. Specifically, the predictions belonging to different class clusters may be close to each other. To solve this, a loss to constrain the relationship between predictions is required. Here we choose the classical triplet loss, which may not be the best, but it works well. Thus $\mathcal{L}_{phrase}$ is defined as:
\begin{equation}
    \mathcal{L}_{phrase} = \mathcal{L}_{distilling} + \beta *\mathcal{L}_{triplet}
    \label{eq:loss_phrase}
\end{equation}
where $\mathcal{L}_{distilling}$ denotes the knowledge distilling function, and we set it to L1 loss, details are shown in ablation study. $\beta$ is a hyper-parameter to balance the weight between distilling loss and triplet loss. The triplet loss $\mathcal{L}_{triplet}$ is defined as:  
\begin{equation}
    \mathcal{L}_{triplet} = max((d(A, P)-d(A, N)+m), 0)
    \label{eq:triplet_loss}
\end{equation}
where $A$ denotes the anchor point which is each output embedding itself, $P$ denotes positive point if it belongs to the same class cluster as $A$, otherwise the point is regarded as negative and denoted as $N$. The notation $d$ is a distance metric, and hyper-parameter $m$ stands for the distance margin.

\subsection{Inference}
The proposed network can parallelly output a group of $N$ HOI instances. We provide two different ways: 1) predict HOI instance from HOI branch, 2) predict HOI instance from phrase branch. This can be formalized as:
\begin{equation}
    \begin{cases}
    C_{hoi} = softmax(O_{hoi})\\
    C_{phrase} = LUT(O_{phrase})
    \end{cases}
\end{equation}

For HOI branch, $C_{hoi}$ denotes the human, object and interaction confidences, which are obtained by applying $softmax$ on the corresponding output $O_{hoi}$ of HOI branch.

For phrase branch, the output embedding $O_{phrase}$ is sent to query its top similar embedding in a given Look-Up-Table, and $C_{phrase}$ is their similarity. The LUT here is often a close-set dictionary, which only contains the embeddings occurred in training set. In this way, the phrase embedding can be translated to an HOI instance. For example, if the top similar embedding in LUT stands for `person sit at dining table', then it can be translated to the HOI instance (human, sit\_at, dining\_table). For both branches, we use shared human and object boxes.

It's noteworthy that there is no need to inference on both branches. In the real-world application, one can remove any one of the branches for memory or computation efficiency. We only report the result of HOI branch in the following experiments if no extra descriptions.
\section{Experiments}

\subsection{Experimental Settings}
\textbf{Datasets. }Experiments are conducted on V-COCO~\cite{datavcoco} and HICO-DET~\cite{datahicodet} benchmark. 
V-COCO is a subset of MS-COCO, which consists of 5,400 images in the trainval dataset and 4946 images in test set. Each human is annotated with binary labels for 29 different action categories. HICO-DET consists of 47,776 images, 38,118 images in training set and 9,658 in test set. It has 600 HOI categories over 117 interactions and 80 objects. 

\textbf{Evaluation Metric. }Following the standard evaluation scheme, we use mean average precision (mAP) to examine the model for both datasets. An HOI detection is considered as true positive if and only if it localizes the human and object accurately, i.e. the Interaction-over-Union (IOU) between the predicted box and ground-truth is greater than 0.5, and predicts the interaction correctly.

\begin{table*}[t]\centering
	
	
	\subfloat[Effectiveness of multi-task.  \label{tab:multi_task}]{
        \setlength{\tabcolsep}{0.7mm}
		\begin{tabular}{l|ccc|ccc}
			\hline
			\multicolumn{1}{l}{} & \multicolumn{3}{|c}{HOI branch} &           \multicolumn{3}{|c}{Phrase branch}  \\
             Methods & Full & Rare & NRare & Full & Rare & NRare \\ 
			\hline 
			HOI model & 23.46 & 16.91 & 25.41 &  - & - & - \\
			HOI model* & 24.07 & 17.01 & 26.18 &  - & - & - \\	
			Phr. model & - &-  & - & 20.09  & 9.96 & 23.12 \\
			PhraseHOI & \textbf{ 29.29}  &\textbf{ 22.03}  & \textbf{31.46}  & 24.40  & 14.04 & 27.50\\
			\hline
		\end{tabular}}\hspace{0.1mm}
	\subfloat[$t_{sim}$ for label composition.  \label{tab:t_sim}]{
    \setlength{\tabcolsep}{1.5mm}
	\begin{tabular}{lccc}
		\hline
		 $t_{sim}$ & Full & Rare & NRare \\
		\hline 
      1.0&27.36  &19.78  &29.63 \\
		0.8 & 28.47  &  21.78 & 30.47 \\
		0.7 &\textbf{29.29 } &\textbf{22.03}  &\textbf{31.46} \\
		0.6 & 27.80&20.52 &29.97 \\
		\hline 
	\end{tabular}}\hspace{0.5mm}
    \subfloat[loss for phrase learning.  \label{tab:distilling_loss}]{
        \setlength{\tabcolsep}{0.8mm}
		\begin{tabular}{lccc}
			\hline
			$loss_{phrase}$ &Full & Rare & NRare \\
			\hline 
			w/o  & 24.07& 17.01&26.18  \\
			\hline
			MSE  & 27.06 & 20.18&29.11    \\
			KL div.  & 26.52 & 18.04 & 29.05\\
			Cosine  & 26.32 & 17.29 & 29.02\\	
			$\mathcal{L}_{distilling}$   & 28.10 & 18.64 & 30.92\\
			\hline
			$\mathcal{L}_{phrase} $  & \textbf{29.29}& \textbf{22.03}&\textbf{ 31.46} \\	
			\hline
		\end{tabular}}\hspace{5mm}
	\caption{Ablations on HICO-DET. (a-c) prove the effectiveness of joint optimization, label composition and phrase learning loss respectively. * denotes the reproduction of ~\cite{zou2021_hoitrans}, which is our baseline. NRare is short for NonRare. }
	\label{tab:ablations}
\end{table*}



\begin{table}[htbp]
\resizebox{\columnwidth}{!}
{
\begin{tabular}{llccc}
\hline \hline
Methods & Backbone  & $AP_{role}$ \\ \hline
\textit{Two-stage methods} \\
VCL~\cite{eccv2020visualcomlear} & R101 &  48.3 \\
Zhou et al.~\cite{zhou2020cascaded} & R50  & 48.9 \\
PastaNet~\cite{li2020pastanet} & R50 & 51.0 \\
DRG~\cite{eccv2020dualgraph} & R50-FPN & 51.0 \\
VSGNet~\cite{ulutan2020vsgnet} & R152 & 51.8 \\
PD-Net~\cite{zhong2020polysemy} & R152   & 52.6 \\ 
ACP~\cite{eccv2020actioncoprior} & R152 & 53.2 \\
\hline \hline
\textit{One-stage methods} \\
UnionDet~\cite{eccv2020uniondet} & R50-FPN  & 47.5 \\
IPNet~\cite{wang2020irnet} & HG-104 & 51.0 \\
AS-NET~\cite{ASNET}  & R50 & 53.9\\
 \hline \hline
\textit{End-to-End methods} \\
HOTR~\cite{HOTR} & R50    & 55.2 \\
QPIC~\cite{qpic2021}  & R50  & 58.8\\
\hline
HOI transformer~\cite{zou2021_hoitrans}    & R50   & 51.2 \\ 
Ours  & R50  & 57.4 \\ 

\hline \hline
\end{tabular}}
\caption{Comparison to state-of-the-art on V-COCO test set. Our method achieves the best on $AP_{role}$.}
\label{tab:vcoco}
\end{table}

\textbf{Implementation Details. }The data augmentation follows~\cite{zou2021_hoitrans}, but the main difference is that, when using scale augmentation, we scale the input image such that the shortest side is at least 448 and at most 640 while the longest at most 900, which is much smaller, resulting in a larger training batch size and a shorter iteration period. This difference also contributes to a little performance gain compared to original HOI Transformer.

The experiments are conducted on two backbones, ResNet-50 and ResNet-101. We use a COCO pre-trained DETR~\cite{detr} to initialize the weights. The model is trained with AdamW, and the learning rate is set to 1e-4 except that the learning rate for backbone is set to 1e-5. The batch size for ResNet-50 and ResNet-101 are set to 64 and 32 respectively, which is 4x larger than that in HOI Transformer. All the models are trained for 200 epochs with once learning rate decay at epoch 150.

The pre-trained word embedding model and language model we use are word2vec~\cite{mikolov2013distributed} and GPT-1~\cite{radford2018improving}, respectively. The hyper-parameter $\alpha$ in Eq.~\ref{eq:loss_total} implies the loss weight of phrase and is set to 0.1, the hyper-parameter $\beta$ in Eq.~\ref{eq:loss_phrase} implies the loss weight of triplet loss and is set to 10, the hyper-parameter $m$ in Eq.~\ref{eq:triplet_loss} is set to 0.5 in experiments.


\subsection{Comparison to State-of-the-Art}
We report quantitative results in terms of $AP$ on HICO-DET in Tab.~\ref{tab:hico} and $AP_{role}$ on V-COCO in Tab.~\ref{tab:vcoco}. For HICO-DET dataset, our method outperforms our baseline~\cite{zou2021_hoitrans} in a large margin, and set state-of-the-art on Full and NonRare categories. Compared to our baseline, we achieve 5.83\% gain on Full, especially 5.12\% gain on Rare categories with ResNet-50. For V-COCO dataset, amazing 6.2\% gain is also achieved compared to our baseline.

\subsection{Ablation Study}
The ablation studies are conducted with ResNet-50 backbone models, and all the models are trained for 200 epochs with once learning rate decay at epoch 150.

\textbf{Phrase Learning and Multi-Task. } The novel relational phrase learning task is derived from HOI, whose labels are converted from HOI annotations. In order to evaluate it, we use its output embeddings to query from LUT to get a result of HOI, which can indicate the quality of the learned embeddings, indirectly reflecting the performance. 

We test on three models to prove the effectiveness, and the results are reported in Tab.~\ref{tab:multi_task}. The first model is our baseline, which has only HOI branch, denoted as HOI model*, the second model is relational phrase learning model, which has only phrase branch with paired (human, object) bounding boxes, denoted as Phr. model, the third model has both branches, denoted as PhraseHOI. 

As shown in Tab.~\ref{tab:multi_task}, the Phr. model achieves 20.09\% on Full and surpasses many recent works in Tab.~\ref{tab:hico}, which proves that the relational phrase learning task (with only one phrase branch) has the ability to model the interaction knowledge. Our baseline is HOI model*, a reproduction of HOI Transformer, which is slightly better than the results reported in~\cite{zou2021_hoitrans}, probably due to the larger training batch size. Compared to the baseline, PhraseHOI achieves 5.22\% gain on Full on HOI branch, which proves the hypothesis that joint optimization may help to distill knowledge from the relational phrase learning task to HOI task. Moreover, compared to Phr. model, PhraseHOI also achieves 4.31\% gain on Full on phrase branch, implying that joint training of the multi-task is beneficial to both of the related tasks.

\textbf{Label Composition. }According to previous section, the proposed novel label composition strategy has two degrees of freedom, one is verb, and the other is object. Best result is obtained when $p_v$ is set to 0.8 and $p_o$ is set to 0.2 in experiments. Here we study different values for the similarity threshold $t_{sim}$ for object and verb neighborhood set. As shown in Tab.~\ref{tab:ablations}(b), $t_{sim}$ = 1.0 means that the neighborhood set has only itself. It can be seen that when setting to 0.8 or 0.7, dramatic improvement are acquired, which indicates that the proposed composition method, based on semantic similarity and manipulated in label space, is effective. However, when setting to 0.6, the performance damages slightly, which is probably caused by the low quality of the neighborhood set when the similarity threshold is getting lower. 

\textbf{Phrase Learning Loss.} Five different losses for phrase embedding learning are studied: MSE loss, KL divergence, cosine loss, L1($\mathcal{L}_{distilling}$) and  $\mathcal{L}_{phrase}$. The distilling loss used here is to constrain the distribution of predictions equal or similar to the distribution of ground truth. As can be seen in Tab.~\ref{tab:ablations}(c), compared with baseline, the distilling loss is effective and gets significantly better result. However, further studies showed that using distilling losses like L1 alone is not enough, because these losses only consider the relationship between predictions and ground truth. On the contrary, relationship between predictions should also be taken into consideration, especially when the ground truths are close to each other. As shown in Tab.~\ref{tab:ablations}(c), using $\mathcal{L}_{phrase}$ can still help improving the performance further. 

At this point, it is clear that in phrase embedding learning, L1 loss plays a more global role to gather all the predictions together around their corresponding  ground truth. Meanwhile, the triplet loss does some subtle tuning in local area to push the similar but category-different points away from each other. It can also be interpreted this way: the L1 loss tries to make the intra-class variance smaller, while the triplet loss helps to make the inter-class variance larger. Results for various $m$ are shown in Tab.~\ref{tab:triplet_loss_margin}. Setting $m=0.50$ gives gains of 1.19\% AP on Full and especial 3.39\% AP on Rare.




\begin{table}[t]
		\centering	
		\setlength{\tabcolsep}{5.8mm}
		\begin{tabular}{lccc}
			\hline
			 $m$ & Full & Rare & NonRare \\
			\hline 
			w/o  & 28.10  &  18.64 & 30.92\\
			\hline
			0.01 & 28.76 & 20.09 & 31.35\\
			0.10 & 29.10 & 21.94 & 31.24\\
			\textbf{0.50} & \textbf{29.29} & \textbf{22.03} & \textbf{31.46}\\
			1.00 & 28.59 & 21.58 & 30.68\\
			\hline 
		\end{tabular}
		\caption{Varying $m$ for triplet loss.}
		\label{tab:triplet_loss_margin}%
\end{table}%

\begin{table}[t]
\setlength{\tabcolsep}{3.8mm}
		\centering	
		\begin{tabular}{lccc}
			\hline
			 Methods & Full & Rare & NonRare \\
			\hline 
			Concatenation  & 29.29 & 22.03 & 31.46 \\	
			LanguageModel  & 22.64 & 14.12 & 25.19 \\
			\hline
		\end{tabular}
		\caption{Different embedding generation methods.}
		\label{tab:embedding_generation}%
		\vspace{-0.3cm}
	\end{table}%
	
\begin{table}[t]
        \setlength{\tabcolsep}{5.1mm}
		\centering	
		\begin{tabular}{lccc}
			\hline
			 Methods & Full & Rare & NonRare \\
			\hline 
			word2vec  & 27.36 & 19.78 & 29.63 \\
			random  & 21.90 & 13.87 & 24.30 \\
			\hline
		\end{tabular}
		\caption{Random word embedding experiment. }
		\label{tab:random_embedding}%
		\vspace{-0.3cm}
	\end{table}%

\textbf{Embedding Generation. }We conduct experiments to test two different embedding generation methods, one is concatenated embedding as shown in Fig~\ref{fig:embedding_generation}(a), the other is language model encoded hidden embedding as shown in Fig~\ref{fig:embedding_generation}(b). The concatenated embedding is the concatenation of three embeddings in `human-verb-object' order, each embedding is queried from word2vec~\cite{le2014distributed} by the query word. If it is a compound word, e.g. `sit at', which can not be found in word2vec, we calculate the mean of each of its separate part. 

As shown in Tab.~\ref{tab:embedding_generation}, 
the concatenation method performs much better in our setting. Further investigations find that the similarities between language model generated embeddings are much higher than that of the concatenated embeddings, e.g. the similarity between `human sip wine glass' and `human toast wine glass' is 0.96 for language model generated embeddings, while only 0.71 for the concatenated embeddings. This may be the reason why the performance of language model generated embedding is poor, because it damages the discriminative ability of knowledge space during training.


\textbf{Analysis. }To further explore the reason why relational phrase helps, we conduct an extra experiment by using a set of random embeddings to replace the word embeddings from word2vec. The result in Tab.~\ref{tab:random_embedding} shows that using random word embeddings is much poorer than that of using word2vec, which implies that it is more the language prior knowledge than the continuous label encoding (compared to discrete one-hot label encoding) structure helps in the multi-task learning.

From an HOI point of view, the proposed PhrasenHOI can be interpreted as knowledge distilling or knowledge transfer from language prior, because inference can be done without the phrase branch, but the performance of HOI branch has been improved.

From the zero-shot view, the proposed PhraseHOI has the potential to do zero-shot prediction, e.g. given an output phrase embedding from phrase branch, if we replace the LUT from close-set to open-set, the output may be semantically reasonable, but this is beyond the scope of this paper.


\section{Conclusion}
In this paper, we propose a derived task from HOI detection, namely relational phrase learning, which encodes semantic knowledge and generates phrases to describe the interactions between humans and objects in image. Then based on the hypothesis that HOI and relational phrase learning may share a common part in their knowledge space, we propose a multi-task architecture named PhraseHOI to jointly optimize them, aiming to improve the discriminative ability and capacity of the shared knowledge space. More efforts are made on the phrase branch, e.g. embedding distilling with triplet loss, label composition to deal with long-tailed problem, which in turn boost the performance of HOI branch. As a result, PhraseHOI gets significant improvement over baseline, setting state-of-the-art on Full and NonRare on the challenging HICO-DET benchmark, which proves the effectiveness of the proposed method. We hope our method can bring new insight to the community.

\textbf{Acknowledgement. } This work is supported by the National Key R\&D Plan of the Ministry of Science and Technology (Project No. 2020AAA0104400) and the special funds from the central finance of Hubei Province guiding the development of local science and technologies grant 2021BEE056. Also, we sincerely thank Bohan Wang and Haoran Wu for their inspirational discussions.

\bibliography{aaai22}

\end{document}